# Universal Adversarial Attack on Deep Learning Based Prognostics

Arghya Basak[1], Pradeep Rathore[1], Sri Harsha Nistala, Sagar Srinivas, Venkataramana Runkana

TCS Research, Pune, 411013, India

Email: {arghya.basak, rathore.pradeep, sriharsha.nistala, sagar.sakhinana, venkat.runkana}@tcs.com

*Abstract*— Deep learning-based time series models are being extensively utilized in engineering and manufacturing industries for process control and optimization, asset monitoring, diagnostic and predictive maintenance. These models have shown great improvement in the prediction of the remaining useful life (RUL) of industrial equipment but suffer from inherent vulnerability to adversarial attacks. These attacks can be easily exploited and can lead to catastrophic failure of critical industrial equipment. In general, different adversarial perturbations are computed for each instance of the input data. This is, however, difficult for the attacker to achieve in real time due to higher computational requirement and lack of uninterrupted access to the input data. Hence, we present the concept of universal adversarial perturbation, a special imperceptible noise to fool regression based RUL prediction models. Attackers can easily utilize universal adversarial perturbations for real-time attack since continuous access to input data and repetitive computation of adversarial perturbations are not a prerequisite for the same. We evaluate the effect of universal adversarial attacks using NASA turbofan engine dataset. We show that addition of universal adversarial perturbation to any instance of the input data increases error in the output predicted by the model. To the best of our knowledge, we are the first to study the effect of the universal adversarial perturbation on time series regression models. We further demonstrate the effect of varying the strength of perturbations on RUL prediction models and found that model accuracy decreases with the increase in perturbation strength of the universal adversarial attack. We also showcase that universal adversarial perturbation can be transferred across different models.

*Keywords*— *Universal adversarial perturbation, Universal adversarial attack, Adversarial attacks on multivariate regression, Adversarial attack on RUL, Adversarial attack on prognostics*

## I. INTRODUCTION

Deep learning models are increasingly being deployed in various mission critical applications such as aircraft structural health monitoring [1], anomaly detection in power plants [2] etc. Several researchers have studied deep learning based RUL prediction for industrial equipment [3]. Industry 4.0 [4] uses state-of-the-art automated Prognostics and Health management (PHM) systems which consist of deep learning based solutions for time series prediction, classification and anomaly detection. These state-of-the-art PHM systems have led to decreased maintenance cost, reduced downtime and more asset availability.Despite tremendous advances in deep learning based PHM systems, very little work has been done to study the security aspects of such systems. Adversarial attacks were first discovered by [5]. Reference [6] proposed the concept of universal adversarial perturbation which could fool deep learning models by adding a special imperceptible noise to any sample in the given dataset. Security aspects of deep learning based time series classification models have been studied extensively in [7] using UCR time series dataset. The authors studied untargeted, targeted and universal adversarial attacks and defenses on deep learning based univariate time series classification models. Recently, [8] studied the effect of adversarial attacks on deep learning based RUL prediction of turbofan engines. These attacks are highly challenging to implement in real time PHM systems as different adversarial perturbations are required to be calculated at each time instance. To compute these adversarial perturbations, an attacker would require continuous access to the input data as well as high computational resources.

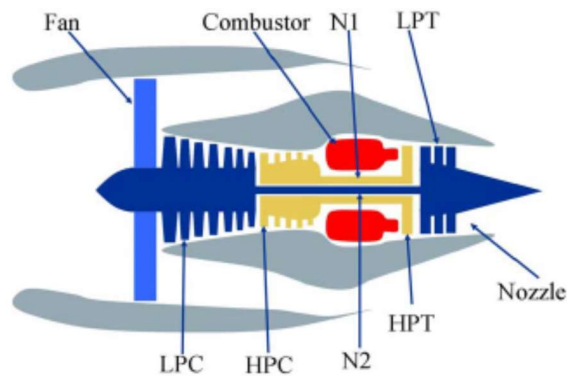

Fig. 1. Illustration of the different subsystems of Turbofan taken from [9]

Universal adversarial perturbation poses much more threat to the security of PHM systems. Only a single adversarial perturbation is required to fool the model at any instance and hence continuous access to the input data by the attacker is not necessary. Also, since a single precomputed universal adversarial perturbation can fool the model at any instant, computational requirement during attack is very low. Because of this destructive capability of universal adversarial perturbation, it is imperative to study the effect of such perturbations on PHM systems. Despite this, no one has studied the vulnerabilities of PHM systems against universal adversarial perturbation so far. In this work, we study the effect of universal adversarial perturbation on RUL prediction for PHM system and demonstrate the same for aircraft turbofan engines. In Fig. 1 [9], a schematic diagram of turbofan engine is shown. A turbofan engine is a complex equipment with interconnected components and is used for propulsion of aircraft. It consists of several subsystems such as combustors, fan, low pressure compressor, high pressure compressor etc. Several variables like total temperature at fan inlet, low pressure compressor, high pressure compressor etc. are sensed at different submodules. The high reliability of remaining useful life predictions of turbofan engines is critical for effective aircraft operation and safety. The turbofan engine dataset of NASA [9] is widely used for benchmarking performances of deep learning models for prognostics and health management.

---
[1] represent equal contribution

The objective of our work is two fold. The first objective is to develop a method for computing universal adversarial perturbation for deep learning based remaining useful life prediction models. The other objective is to showcase the transferability of universal adversarial attacks across different models.

The contribution of our work can be summarized as follows:

1) We propose a novel algorithm to compute universal adversarial perturbation for multivariate time series regression based RUL prediction problem.
2) We showcase that universal adversarial perturbation can be used to attack and fool deep learning based PHM systems in real time.
3) We demonstrate that the prediction accuracy of deep learning based RUL model decreases with the increase in perturbation strength of the universal adversarial perturbation.
4) We show the transferability of universal adversarial perturbation across different models i.e. universal adversarial perturbation computed using one model can be utilized to fool another model.

The rest of the paper is organized as follows. In section II, we will explain about the mathematical notations and algorithm for computing universal adversarial perturbations. Section III contains the experimental details of our study. In section IV, we will discuss our results and finally conclude our work.

## II. BACKGROUND

In this study, we have formulated the RUL prediction task as a multivariate time series regression problem. In this section, we introduce the concept of adversarial perturbation and universal adversarial perturbation, followed by the mathematical formulation for the same. Notations used in this study are given below

- $X^i \in R^{MXN}$ is the $i^{th}$ sample of the dataset, $X$ where M and N represents the input sequence length to the model and number of input features respectively
- $Y^i \in R$ denotes the true target value for $i^{th}$ sample
- $U \in R^{MXN}$ represents the universal adversarial perturbation for input samples
- $\varepsilon \in R$ represents the allowable limit of $L_\infty$ norm of the permitted perturbation

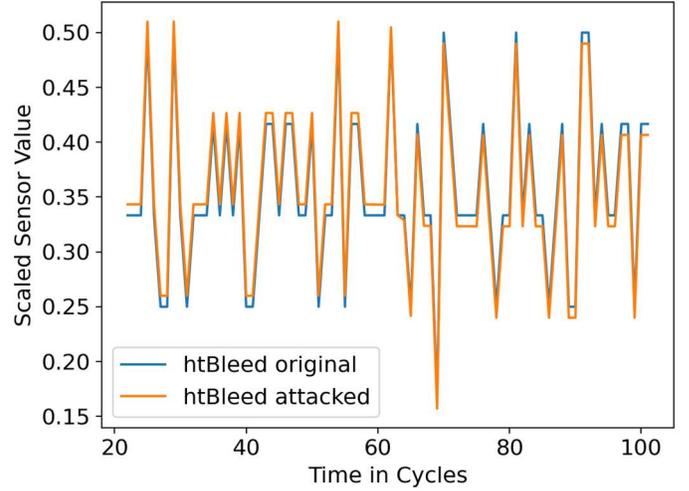

Fig. 2. Original and adversarial sensor data for sensor Bleed Enthalpy (htBleed) for Engine ID 79

- $X_{adv}^i \in R^{MXN}$ represents the adversarial sample corresponding to sample $X^i$
- $f(.): R^{MXN} \to R$ is a neural network model which outputs the target value
- $\hat{Y}^i$ is the predicted target value by $f(.)$ for $X^i$
- $L(Y^i, \hat{Y}^i)$ is the regression loss function quantifying the error between the true and predicted target value corresponding to input sample $X^i$
- $X_{FGSM,\varepsilon}^i \in R^{MXN}$ represents the adversarial sample corresponding to $X^i$ using Fast Gradient Sign Method (FGSM)
- $R_{fool} \in R$ is the fooling ratio representing the desired fraction of samples to be fooled by universal adversarial perturbation
- $N_{test}$ is the number of test samples
- $E_{fool}$ denotes the maximum number of epochs for calculating universal adversarial perturbation
- $check_{fool}(X + U)$ is the fooling ratio of the dataset X after adding universal perturbation, U to each sample

Reference [10] showed that deep learning based models are susceptible to adversarial attacks. The authors proposed Fast Gradient Sign Method (FGSM) algorithm to find adversarial perturbation for any sample in a single iteration. In FGSM, the adversarial perturbation is determined by taking a step in the direction of the 'sign of gradient of loss' with respect to input samples. Mathematically,

$$X_{FGSM,\varepsilon}^i = X^i + \varepsilon . sign\left( \nabla_{X^i} L(Y^i, \hat{Y}^i) \right) \quad (1)$$

Adversarial attacks were able to fool deep learning based time series classification models for univariate datasets [11]. Reference [12] successfully performed FGSM attacks on deep learning based time series classification models for multivariate datasets. There exists universal adversarial perturbation [6] that can be added to any image of a given dataset to fool the deep

**Algorithm 1**: Computing universal adversarial perturbation

**Input:** $X, Y, f, \varepsilon, R_{fool}, E_{fool}$
**Output:** $U$
1: $U \leftarrow 0$
2: $Iter \leftarrow 0$
3: **while** $check_{fool}(X + U) < R_{fool}$ And $Iter < E_{fool}$ **do**
4:    **for** $X^i \in X$ **do**
5:      **if** $f(X^i + U) < (1 + \alpha) Y^i$
6:       **Solve for minimum $r$ in the direction of gradient of loss w.r.t.** $(X^i + U)$ **i.e.** $\nabla_{(X^i+U)} L(Y^i, f(X^i + U))$
        $\Delta U = \underset{r}{\arg\min} \|r\|$ s.t. $f(X^i + U + r) > (1 + \alpha) Y^i$
7:       Take projection of $U + \Delta U$ on $\varepsilon$ infinity ball
        $U \leftarrow Projection_\varepsilon (U + \Delta U)$
8:      **end if**
9:    **end for**
10:   $Iter += 1$
11:   Shuffle $X$
12: **end while**

learning based image classification model. Universal adversarial attacks were also performed successfully on univariate time series classification models in [7]. In the present work, we tested the vulnerability of multivariate time series regression based prognostic models to universal adversarial attacks. Figure 2 shows an example of original sensor data and sensor data after adding adversarial perturbation. The example shows that the original sample and the adversarial sample are almost indistinguishable.

*A. Universal Adversarial Perturbation*

In this subsection, we propose a method to find universal adversarial perturbation for deep learning based multivariate time series regression models. Universal adversarial perturbation increases the error in prediction of the target value corresponding to majority of the samples selected from the input data distribution, $\rho$. Mathematically,

$$| f(X^i + U) - f(X^i) | \geq \alpha \ for \ X^i \sim \rho \qquad (2)$$

Here $\alpha$ represent the acceptable threshold for error corresponding to deep learning based time series regression model and its value depends on the use case. In the case of RUL prediction for industrial equipment, both overprediction and underprediction are undesirable. However, overprediction is much more dangerous as it can lead to insufficient planning of scheduled maintenance leading to catastrophic failure of critical industrial equipment. It can cost human lives and incur huge capital loss due to damage of critical infrastructure. On the other hand, underprediction of RUL is less catastrophic as it only leads to additional maintenance cost due to early scheduling of maintenance activities. Hence, in this work we focus on the problem of overprediction of RUL of critical industrial equipment due to universal adversarial attack. We have utilized NASA's Turbofan engine dataset [9] for our study as turbofan engine is the most critical part of an aircraft.

*B. Adversarial Attack on Turbofan*

In this work, we showcase the vulnerability of deep learning based time series models for overprediction of RUL via universal adversarial attacks. Algorithm 1 represents the method to find such universal adversarial perturbations which can lead to overprediction of turbofan RUL. Universal adversarial perturbations should follow two criteria. Firstly, the $L_\infty$ norm of the universal adversarial perturbations should be less than $\varepsilon$ Secondly, universal adversarial perturbation should lead to overprediction of at least $R_{fool}$ fraction of samples by a factor of $\alpha$. Mathematically the two criteria may be expressed as,

$$\|U\|\infty \leq \varepsilon \qquad (2)$$

$$\underset{X^i \sim \rho}{\mathrm{P}} ( f(X^i + U) > (1 + \alpha)Y^i ) \geq R_{fool} \qquad (3)$$

Here, P represent the probability of computing the successful attacks. Kindly note that there exist multiple universal adversarial perturbations for a given dataset which can be obtained by random shuffling of the dataset used to compute universal adversarial perturbation.

### III. EXPERIMENTAL DETAILS

In this study we have used multivariate time series data generated for turbofan engines via thermodynamic simulations using an aero-propulsion simulator, C-MAPSS [13]. Time series data are generated for different sensors of the aircraft engine. The dataset FD001 is used for training and testing the deep learning based time series regression model. FD001 has 100 aircraft engines data for training and another separate 100 aircraft data for test. The faults were introduced artificially during the engine simulation and were allowed to evolve till failure criteria are met. Data from 21 sensors is provided till failure. Out of these 21 variables, we selected 14 variables as the remaining 7 variables were constant. We normalized all the sensor data between 0 and 1. The details of the different model architectures used to predict RUL can be found in [8].

We have used multiple deep learning based multivariate time series regression models by leveraging LSTM [14] and GRU [15]. The length of the time window was taken to be 80 for all the models. We also tried different values of the parameter, $\varepsilon$ representing strength of universal adversarial perturbations and studied their effect on model accuracy. Note that adversarial perturbations are univariate and multivariate in case of univariate and multivariate time series data respectively.

### IV. RESULTS AND DISCUSSIONS

In this section, we discuss about the results obtained from our study on universal adversarial attacks on deep learning based RUL prediction models. We used two different deep learning models namely LSTM and GRU for RUL prediction for turbofan engines. We used our novel algorithm to generate universal adversarial perturbation for both models and studied their effect on prediction accuracy of the models. We also studied the transferability property of universal adversarial perturbations generated using one model and investigated their efficacy on the other model. The effect of varying the strength of universal adversarial perturbation on the performance of the models is also studied.

*A. Computing Universal Adversarial Perturbation*

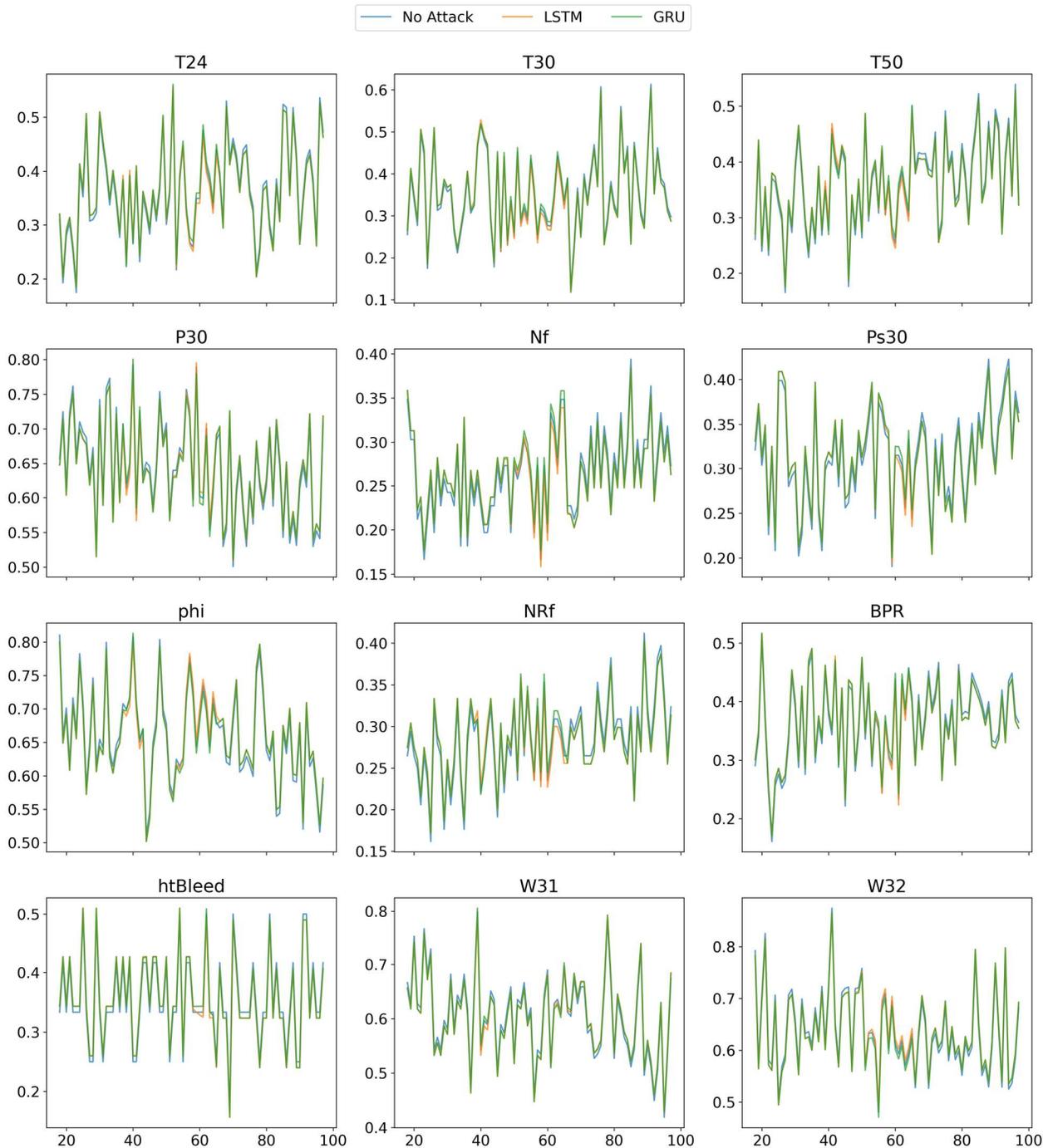

Fig. 3. Time series plot of sensor values from Engine ID 79 for original sample and corresponding adversarial samples computed using LSTM and GRU models with ε of 0.01. X-axis represents time in unit of cycles

We used Algorithm 1 to generate universal adversarial perturbations. Fig. 3 shows a multivariate time series sample chosen from data corresponding to Engine ID 79 in the test dataset. As shown in Fig 3, adversarial samples generated using LSTM and GRU models are very similar to the original sample for every sensor and are visually indistinguishable from the original sample. Out of 14 sensors used as input to the LSTM and GRU models, 12 sensors are plotted in Fig 3. We chose perturbation strength $\varepsilon$, overprediction factor $\alpha$, target fooling ratio $R_{fool}$ and number of epochs $E_{fool}$ as 0.01, 0.1, 0.99 and 3 respectively. Our universal adversarial attack algorithm focuses specifically on overprediction of RUL. The effect of adversarial perturbation strength on RUL prediction is discussed in the next section.

## B. Effect of universal adversarial perturbation strength

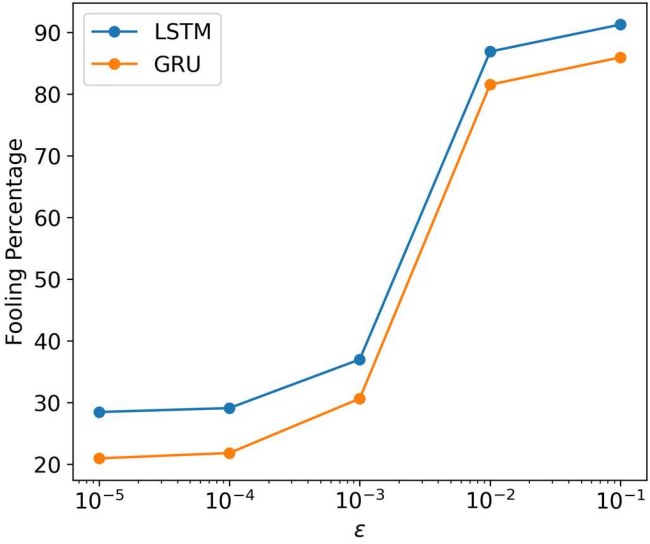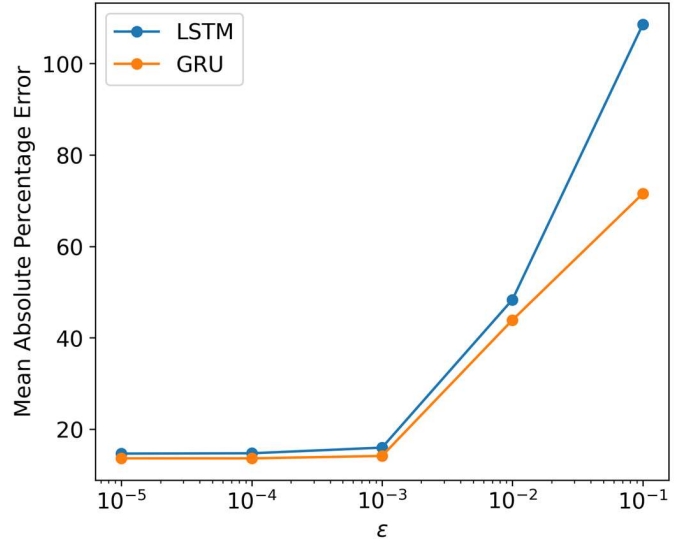

Fig. 4. Effect of universal adversarial perturbation strength on various model metrics

We did extensive experimentation to determine the optimum value of ε such that the universal adversarial perturbation remains imperceptible while still being able to fool the model with significant drop in model performance. We varied ε on log scale to obtain an appropriate order of magnitude. Two metrics, fooling percentage and mean absolute percentage error (MAPE) as described mathematically below, are chosen to investigate the effect of changing strength of universal adversarial perturbation

$$Fooling\ Percentage = \frac{No.\ of\ fooled\ samples}{No.\ of\ test\ samples} X\ 100 \quad (4)$$

$$MAPE\quad \frac{1}{N_{test}}\sum_{i=1}^{N_{test}} \frac{|\hat{Y}^i - Y^i|}{Y^i} X100 \quad (5)$$

Fig. 4 shows the effect of changing universal adversarial perturbation strength, ε on fooling percentage and MAPE of RUL prediction using LSTM and GRU models. As shown in Fig 4, fooling percentage as well as MAPE on test data increase with the increase in ε. For very low value of ε , the perturbation is too small to make any significant change in the input data and hence there is no significant change in model performance metrics. Large values of ε achieve high fooling ratios and MAPE on test data but the universal adversarial perturbation does not remain imperceptible. Hence, we finally choose the optimum value of ε to be 0.01 as the corresponding adversarial perturbation remains visually imperceptible while being able to achieve high fooling ratio and MAPE. Imperceptibility of universal adversarial perturbation corresponding to ε equal to 0.01 is also demonstrated in Fig 3. The effect of variation in ε on model metrics is similar for both LSTM and GRU based RUL prediction models as depicted in Fig 4. Specifically, for ε equal to 0.01, the difference in fooling percentage and MAPE of LSTM and GRU is very small.

## C. Comparision of predicted RUL before and after attack

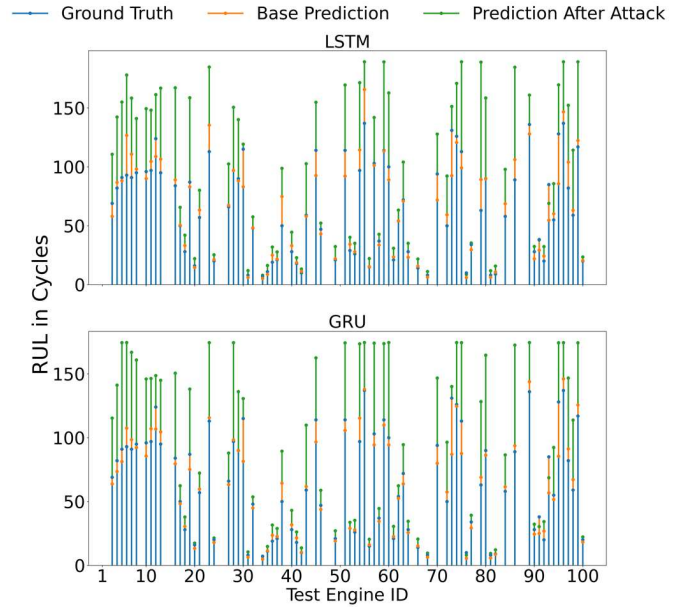

Fig. 5. RUL prediction before and after attack for engines in test dataset

Fig. 5 shows the comparison of predicted RUL for the last time window of each engine in the test dataset before and after universal adversarial attack. For majority of engines, universal adversarial perturbation is able to overpredict the RUL by significant margin for both LSTM as well as GRU models.

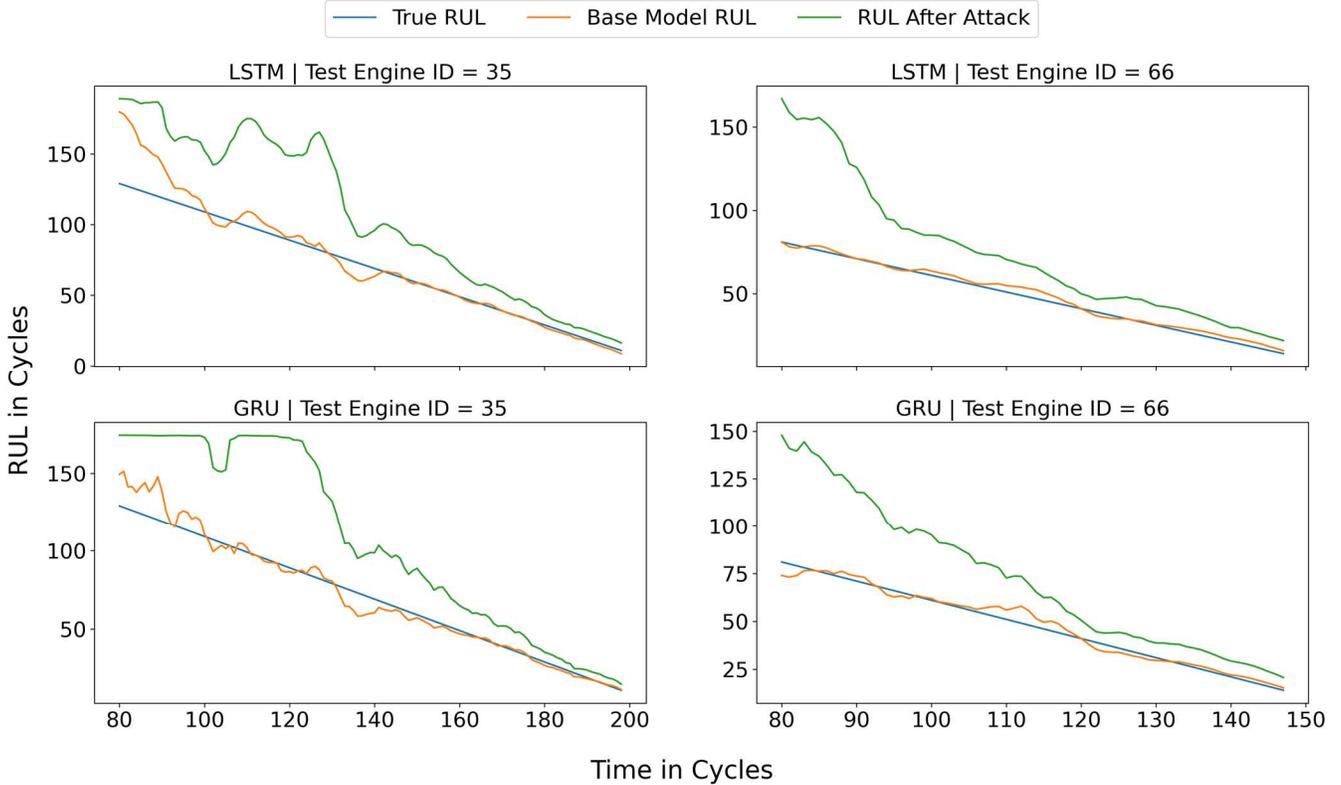

Fig. 6. Comparison of Time Series plots of true and predicted RUL

This shows the effectiveness of universal adversarial attacks on different RUL prediction models for the entire dataset. Fig. 6 compares RUL predictions for two engines before and after attack over time. During the initial cycles, the error between true RUL and the RUL predicted using LSTM and GRU baseline models is higher because RUL depends on the usage history and trajectory of operating conditions of the engines during operation. Hence, accurate long horizon prediction of RUL is not possible in general. Over the course of engine operation, RUL prediction using the models became better as operating history of the engines is able to capture the engine health trajectory fairly well. As shown in Fig. 6, both LSTM and GRU models significantly overpredict RUL compared to both baseline model prediction and true RUL after universal adversarial attack. These overpredictions of RUL can lead to postponement of planned maintenance activities and can cause abrupt failure of the engine. It should also be noted that even small overprediction towards the end of engine life can be very disastrous because it may lead to lack of preparedness for engine maintenance and sudden unexpected failure of the aircraft engine. Due to this fact, we have chosen relative percentage overprediction (10% above true value) as a criterion for the success of universal adversarial attack.

D. *Transferability of universal adversarial perturbation*

In this section, we studied the transferability property of universal adversarial perturbation across both LSTM and GRU based RUL prediction models. Transferability of adversarial perturbation means that adversarial perturbation computed using a deep learning model can be utilized to misguide or fool another deep learning model having same or different architecture. Table 1. shows the performances of both LSTM and GRU based RUL prediction models against universal adversarial perturbations generated using either LSTM or GRU model. Note that fooling percentage in case of no attack is representing the performance of base models. As shown in Table 1, the fooling percentage and MAPE on test data for both LSTM and GRU based RUL prediction models increase significantly when adversarial perturbation is added to the test data. Fooling percentage and MAPE of LSTM based RUL prediction model increase from 28.41 and 14.66 to 85.05 and 46.69 respectively when universal adversarial perturbation computed using the GRU model is added to the test data.

Table 1. Performance of RUL prediction models on test data against attack generated using LSTM and GRU

| RUL Prediction Model | Attack Model | Fooling Percentage (%) | MAPE |
|---|---|---|---|
| LSTM | None | 28.41 | 14.66 |
| LSTM | LSTM | 86.86 | 48.33 |
| LSTM | GRU | 85.05 | 46.69 |
| GRU | None | 20.87 | 13.63 |
| GRU | GRU | 81.50 | 43.88 |
| GRU | LSTM | 81.27 | 44.15 |

Similarly, for other scenarios results are given in Table 1. It can be observed from Table 1 that irrespective of the model used for computing universal adversarial perturbation, the effect of these perturbations on performance of LSTM model as well as GRU model is very similar. This demonstrates the successful transferability of these perturbations.

Transferability of adversarial perturbation increases the threat to industrial AI systems because such perturbations can misguide or fool AI models even when they are not accessible to the attacker. In real life scenarios, the attacker may use his own surrogate model(s) to generate universal adversarial perturbations and utilize them to perform attacks on the actual models being used in the plant, process or equipment.

*E. Conclusion*

In this work, we studied universal adversarial attacks on deep learning based multivariate time series prediction models. We proposed a novel algorithm for computing universal adversarial perturbation for deep learning based time series prediction models and the effect of such perturbations are studied on deep learning based RUL prediction models using the NASA turbofan dataset. We find that the imperceptibility and transferability of these universal adversarial perturbations pose severe threat to the modern deep learning based AI solutions for industrial systems such as aircraft engine. In our future work, we would like to develop algorithms to increase the robustness of deep learning based time series models against adversarial attacks.

ACKNOWLEDGMENT

We would like to thank the authors of "Crafting Adversarial Examples for Deep Learning Based Prognostics" and "Damage propagation modeling for aircraft engine run-to-failure simulation" for providing their pretrained deep learning models and dataset respectively. We utilized the LSTM and GRU based pretrained models and the dataset provided by the authors in our study.